\documentclass[]{article}

\usepackage{colacl}
\usepackage{amsmath}
\usepackage{qtree}

\title{Trainable Methods for Surface Natural Language Generation}
\author{Adwait Ratnaparkhi\\
IBM TJ Watson Research Center\\
P.O. Box 218\\
Yorktown Heights, NY 10598\\
{\tt aratnapa@us.ibm.com}}

\begin{document}

\maketitle

\abstract{
We present three systems for surface natural language generation that
are trainable from annotated corpora.  
The first two systems, called NLG1 and NLG2, require
a corpus marked only with domain-specific semantic attributes, while
the last system, called NLG3, requires a corpus marked with both
semantic attributes and syntactic dependency information.  
All systems attempt to produce
a grammatical natural language phrase from a domain-specific semantic
representation.  
NLG1 serves a baseline system and uses phrase frequencies to generate a whole phrase in one step, 
while NLG2 and NLG3 use maximum entropy probability models to individually
generate each word in the phrase.  The systems NLG2 and NLG3
learn to determine both the word choice and the word order of the phrase.
We present experiments in which we generate phrases to describe flights
in the air travel domain.  }

\newcommand{\stopsym}{{\tt *stop*}}

\section{Introduction}

This paper presents three trainable systems for surface natural language
generation (NLG). Surface NLG, for our purposes, consists of
generating a grammatical natural language phrase that expresses the
meaning of an input semantic representation.  The systems take a
``corpus-based'' or ``machine-learning'' approach to surface NLG, and
learn to generate phrases from semantic input by statistically
analyzing examples of phrases and their corresponding semantic
representations.  The determination of the content in the semantic
representation, or ``deep'' generation, is not discussed here.
Instead, the systems assume that the input semantic representation is
fixed and only deal with how to express it in natural language.

This paper discusses previous approaches to surface NLG,
and introduces three trainable systems for surface NLG, called NLG1, NLG2, and NLG3.
Quantitative evaluation of experiments in the air travel domain will also be discussed.

\section{Previous Approaches}

Templates are the easiest way to implement surface NLG. A template for
describing a flight noun phrase in the air travel domain might be
\begin{tt}
flight departing from \$city-fr at \$time-dep and arriving in \$city-to
at \$time-arr
\end{tt}
where the words starting with ``{\tt \$}'' are actually variables
---representing the departure city, and departure time, the arrival
city, and the arrival time, respectively--- whose values will be
extracted from the environment in which the template is used.  The
approach of writing individual templates is convenient, but may not
scale to complex domains in which hundreds or thousands of templates
would be necessary, and may have shortcomings in maintainability and
text quality (e.g., see \cite{templatesvsnlg} for a discussion).

There are more sophisticated surface generation packages,
such as FUF/SURGE \cite{surge96}, KPML \cite{kpml:tech}, MUMBLE \cite{mumble}, 
and RealPro \cite{realpro}, which produce natural language text 
from an abstract semantic representation. 
These packages require linguistic sophistication in order
to write the abstract semantic representation, but they 
are flexible because minor changes to the input can 
accomplish major changes to the generated text.

The only trainable approaches (known to the author) to surface
generation are the purely statistical machine translation (MT) systems 
such as \cite{maxent:mt} and the corpus-based generation
system described in \cite{langkilde98}.  The MT systems of
\cite{maxent:mt} learn to generate text in the target language
straight from the source language, without the aid of an explicit
semantic representation.  In contrast, \cite{langkilde98} uses
corpus-derived statistical knowledge to rank plausible hypotheses 
from a grammar-based surface generation component.

\section{Trainable Surface NLG}

%
% motivate trainable NLG
%
In trainable surface NLG, the goal is to learn
the mapping from semantics to words that would
otherwise need to be specified in a grammar
or knowledge base. 
All systems in this paper use  {\em attribute-value} pairs
as a semantic representation, which suffice as a representation
for a limited domain like air travel.
For example, the set of attribute-value pairs
\{ \$city-fr = New York City, \$city-to = Seattle , 
\$time-dep = 6 a.m., \$date-dep = Wednesday \} 
represent the meaning of the noun phrase
``a flight to Seattle that departs from New York City at 6 a.m. on Wednesday''.
The goal, more specifically, is then to learn
the optimal {\em attribute ordering} and {\em lexical choice}
for the text to be generated from the attribute-value pairs.
For example, the NLG system should automatically decide
if the attribute ordering in  ``flights to New York in the evening''
is better or worse than the ordering in 
``flights in the evening to New York''. 
Furthermore, it should automatically decide if 
the lexical choice in 
``flights departing to New York'' is better or worse
than the choice in ``flights leaving to New York''. 
The motivation for a trainable surface generator is
to solve the above two problems in a way that reflects
the observed usage of language in a corpus, but without
the manual effort needed to construct a grammar or
knowledge base.

All the trainable NLG systems in this paper
assume the existence of a large corpus 
of phrases in which the values of interest have been
replaced with their corresponding attributes, or in 
other words, a corpus of {\em generation templates}.
Figure~\ref{trdatasample} shows a sample of training data, 
where only words marked with a ``\$'' are attributes.
All of the NLG systems in this paper work in two steps as shown
in Table~\ref{twosteps}.
The systems NLG1, NLG2 and NLG3 all implement step 1; they produce 
a sequence of words intermixed with attributes, i.e., a template, from the
the attributes alone. The values are ignored until step 2, when they replace
their corresponding attributes in the phrase produced by step 1.

\begin{table*}
\begin{verbatim}
flights on $air from $city-fr to $city-to the $time-depint of $date-dep 
$trip flights on $air from $city-fr to $city-to leaving after $time-depaft on $date-dep 
flights leaving from $city-fr going to $city-to after $time-depaft on $date-dep 
flights leaving from $city-fr to $city-to the $time-depint of $date-dep 
$air flight $fltnum from $city-fr to $city-to on $date-dep 
$city-fr to $city-to $air flight $fltnum on the $date-dep 
$trip flights from $city-fr to $city-to 
\end{verbatim}
\caption{Sample training data}
\label{trdatasample}
\end{table*}

\begin{table*}
\begin{tabular}{lp{3.0in}}
Input to Step 1:&\{ \$city-fr, \$city-to, \$time-dep, \$date-dep \} \\
Output of Step 1:&``a flight to \$city-to that departs from \$city-fr at \$time-dep on \$date-dep''\\
\\
Input to Step 2:&``a flight to \$city-to that departs from \$city-fr at \$time-dep on \$date-dep'', \{ \$city-fr = New York City, \$city-to = Seattle , \$time-dep = 6 a.m., \$date-dep = Wednesday \} \\
Output of Step 2:&``a flight to Seattle that departs from New York City at 6 a.m. on Wednesday''\\
\end{tabular}
\caption{Two steps of NLG process}
\label{twosteps}
\end{table*}

\newcommand{\argmax}{\operatornamewithlimits{argmax}}

\subsection{NLG1: the baseline}

The surface generation model NLG1 simply chooses the
most frequent template in the training data that corresponds
to a given set of attributes. 
Its performance is intended to serve as a baseline result to the more 
sophisticated models discussed later.
Specifically, $nlg_1(A)$
returns the phrase that corresponds to the attribute
set $A$: 
\[
nlg_1(A) = 
\left\{ 
\begin{array}{ll}
\argmax_{phrase \in T_{A}} C(phrase, A)&{T_{A} \neq \emptyset}\\
\mbox{[empty string]}&{T_{A} = \emptyset}
\end{array}
\right. \]
where $T_{A}$ are the phrases that have occurred with $A$ in the 
training data, and where $C(phrase, A)$ is the training data 
frequency of the natural language phrase $phrase$ and the set of attributes $A$. 
NLG1 will fail to generate anything if $A$ is a novel combination of attributes.

\subsection{NLG2: $n$-gram model}

The surface generation system NLG2 assumes that the best choice to
express any given attribute-value set is the word sequence with the
highest probability that mentions all of the input attributes exactly once.
When generating a word, it uses local information, captured by word
$n$-grams, together with certain non-local information, namely, the
subset of the original attributes that remain to be generated.  The
local and non-local information is integrated with use of features in
a maximum entropy probability model, and a highly pruned search
procedure attempts to find the best scoring word sequence according to
the model.

\subsubsection{Probability Model}

\newcommand{\nlgtwohist}{w_{i-1}, w_{i-2}, attr_i}

The probability model in NLG2 is a conditional distribution over
$V \cup \stopsym$, where $V$ is the generation vocabulary and 
where $\stopsym$ is a special ``stop'' symbol. 
The generation vocabulary $V$ consists of all the words 
seen in the training data.
The form of the maximum entropy probability model 
is identical to the one used in \cite{maxent:mt,ratnaparkhi:thesis}:
\begin{small}
\begin{eqnarray*}
p(w_i|\nlgtwohist) &=& {{\prod_{j=1}^k \alpha_j^{f_j(w_i,\nlgtwohist)} } \over Z(\nlgtwohist)} \\ 
Z(\nlgtwohist) &=& \sum_{w'} \prod_{j=1}^k \alpha_j^{f_j(w',\nlgtwohist)} 
\end{eqnarray*}
\end{small}
where $w_i$ ranges over $V \cup \stopsym$ and
$\{ \nlgtwohist \}$ is the history, where $w_i$ denotes the $i$th word in 
the phrase, and $attr_i$ denotes the attributes that remain to be
generated at position $i$ in the phrase.
The $f_j$, where  $f_j(a,b) \in \{0,1\}$,  are called {\em features}
and capture any information in the history that might be useful
for estimating $p(w_i|\nlgtwohist)$.
The features used in NLG2 are described in the next section, 
and the feature weights $\alpha_j$, obtained from
the Improved Iterative Scaling algorithm \cite{maxent:mt}, 
are set to maximize the likelihood of the training data.
The probability of the sequence $W=w_1 \dots w_n$, 
given the attribute set $A$, (and also given that its length is $n$) is:
\begin{eqnarray*}
Pr(W=w_1 \dots w_n | len(W)=n, A) &=&\\
\prod_{i=1}^n p(w_i | \nlgtwohist)&&\\ 
\end{eqnarray*}

\subsubsection{Feature Selection}

\begin{table*}
\begin{tabular}{lp{4.0in}}
Description&Feature $f(w_i,\nlgtwohist) = \dots$\\ \hline
No Attributes remaining&1 if $w_i=?$ and $attr_i = \{ \empty \}$, 0 otherwise\\ 
Word bi-gram with attribute&1 if $w_i=?$ and $w_{i-1}=?$ and $? \in attr_i$, 0 otherwise\\
Word tri-gram with attribute&1 if $w_i=?$ and $w_{i-1} w_{i-2}=? ?$ and $?  \in attr_i$, 0 otherwise
\end{tabular}
\caption{Features patterns for NLG2. Any occurrence of ``?'' will be instantiated with an actual value from training data.}
\label{nlg2features}
\end{table*}

The feature patterns, used in NLG2 are shown
in Table~\ref{nlg2features}. The actual features are created
by  matching the patterns over the training data, e.g., 
an actual feature derived from the word bi-gram template might be:
\begin{small}
\[ f(w_i,\nlgtwohist) = 
\left\{ 
\begin{array}{ll}
1&\mbox{if $w_i={\tt from}$}\\
&\mbox{and $w_{i-1} = {\tt flight}$} \\
&\mbox{and ${\tt \$city-fr} \in attr_i$} \\  % $ for emacs hilighting
0&\mbox{otherwise}
\end{array}
\right. 
\]
\end{small}
Low frequency features involving word $n-$grams tend to be unreliable; the NLG2 system
therefore only uses features which occur $K$ times or more in the training data.

\subsubsection{Search Procedure}

The search procedure attempts to find a word sequence
$w_1 \dots w_n$ of any length $n \leq M$ for the input attribute set $A$ such that 
\begin{enumerate}
\item $w_n$ is the stop symbol $\stopsym$
\item All of the attributes in $A$ are mentioned at least once
\label{atleastonce}
\item All of the attributes in $A$ are mentioned at most once
\label{atmostonce}
\end{enumerate}
and where $M$ is an heuristically set maximum phrase length.

The search is similar to a left-to-right breadth-first-search, except that
only a fraction of the word sequences are considered.
More specifically, the search procedure implements the recurrence:
\begin{eqnarray*}
W_{N, 1} &=& top(N, \{ w | w \in V \})\\
W_{N,i+1} &=& top(N, next(W_{N,i}))
\end{eqnarray*}
The set $W_{N,i}$ is the top $N$ scoring sequences of length $i$, 
and the expression $next(W_{N,i})$ returns all sequences $w_1 \dots w_{i+1}$
such that $w_1 \dots w_i \in W_{N,i}$, and $w_{i+1} \in V \cup \stopsym$. 
The expression $top(N, next(W_{N,i}))$ finds the top $N$ sequences in 
$next(W_{N,i})$. 
During the search, any sequence that ends with $\stopsym$ is removed
and placed in the set of completed sequences. 
If $N$ completed hypotheses are discovered, or if $W_{N, M}$ is computed, the search terminates.
Any incomplete sequence which does not satisfy condition (\ref{atmostonce})
is discarded and any complete sequence that does not satisfy condition (\ref{atleastonce})
is also discarded. 

When the search terminates, there will be at most $N$ completed sequences, of possibly
differing lengths.
Currently, there is no normalization for different lengths, i.e., 
all sequences of length $n \leq M$ are equiprobable:
\begin{eqnarray*}
Pr(len(W)=n) &= {1 \over M} & n \leq M\\
&=0&n > M\\
\end{eqnarray*}  
NLG2 chooses the best answer to express the attribute set $A$ as follows: 
\begin{eqnarray*}
nlg_2(A) = &\argmax_{W \in W_{nlg2}}& Pr(len(W)=n) \cdot \\
&&Pr( W | len(W)=n, A) 
\end{eqnarray*}
where $W_{nlg2}$ are the completed word sequences that satisfy the conditions
of the NLG2 search described above.

\subsection{NLG3: dependency information}

\begin{figure*}
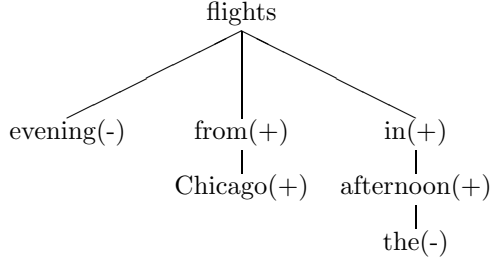

\begin{center}
\Tree [.flights evening(-) [.from(+) Chicago(+) ] [.in(+) [.afternoon(+) the(-) ] ] ]
\end{center}
\caption{Sample dependency tree for the phrase {\em evening flights from Chicago in the afternoon}. - and + signs indicate left or right child, respectively.}
\label{sampledeptree}
\end{figure*}

NLG3 addresses a shortcoming of NLG2, namely that
the previous two words are not necessarily the
best informants when predicting the next word. 
Instead, NLG3 assumes that conditioning on {\em syntactically related}
words in the history will result on more accurate surface generation. 
The search procedure in NLG3 generates a syntactic dependency tree
from top-to-bottom instead of a word sequence from left-to-right, 
where each word is predicted in the context of its syntactically
related parent, grandparent, and siblings.
NLG3 requires a corpus that has been annotated with tree
structure like the sample dependency tree shown in Figure~\ref{sampledeptree}. 

\subsubsection{Probability Model}

\newcommand{\nlgthreehist}{w, ch_{i-1}(w), ch_{i-2}(w), par(w), dir, attr_{w,i}}
\newcommand{\nlgthreehistdir}[1]{w, ch_{i-1}(w), ch_{i-2}(w), par(w), dir=#1, attr_{w,i}}

The probability model for NLG3, shown in Figure~\ref{nlg3childequations}, conditions on the parent, the two closest 
siblings, the direction of the child relative to the parent, and the attributes that remain to be generated.
\begin{figure*}
\begin{small}
\(
\begin{array}{|lll|} \hline
p(ch_i(w)|\nlgthreehist) &=& {\prod_{j=1}^k \alpha_j^{f_j(ch_i(w), \nlgthreehist)} \over Z(\nlgthreehist)} \\ 
&&\\
Z(\nlgthreehist) &=& \sum_{w'} \prod_{j=1}^k \alpha_j^{f_j(w',\nlgthreehist)}  \\ \hline
\end{array}
\)
\end{small}
\caption{NLG3: Equations for the probability of the $i$th child of head word $w$, or $ch_i(w)$}
\label{nlg3childequations}
\end{figure*}

Just as in NLG2, $p$ is a distribution over $V \cup \stopsym$, and the
Improved Iterative Scaling algorithm is used to find the feature
weights $\alpha_j$.  The expression $ch_i(w)$ denotes the $i$th
closest child to the headword $w$, $par(w)$ denotes the parent of
the headword $w$, $dir \in \{ {\tt left}, {\tt right} \}$ denotes
the direction of the child relative to the parent, 
and $attr_{w,i}$ denotes the attributes that remain
to be generated in the tree
when headword $w$ is predicting its $i$th child.
For example, in Figure~\ref{sampledeptree}, if $w=$``flights'', then
$ch_1(w) = $``evening'' when generating the left children, and $ch_1(w) = $``from''
when generating the right children.
As shown in Figure~\ref{nlg3treeequations}, the probability of a dependency tree
that expresses an attribute set $A$ can be found by computing, for each word in the
tree, the probability of generating its left children and then its
right children.\footnote{We use a dummy ROOT node to generate
the top most head word of the phrase}
\begin{figure*}
\begin{small}
\(
\begin{array}{|lll|} \hline
Pr(T | A) &=& \prod_{w \in T} Pr_{left}(w | A) Pr_{right}(w | A) \\
&&\\
Pr_{left}(w | A) &=& Pr(\mbox{\# of left children} = n) 
\prod_{i=1}^n p(ch_i(w) | \nlgthreehistdir{{\tt left}} )\\
&&\\
Pr_{right}(w | A) &=& Pr(\mbox{\# of right children} = n)
\prod_{i=1}^n p(ch_i(w) | \nlgthreehistdir{{\tt right}} )\\ \hline
\end{array}
\)
\end{small}
\caption{NLG3: Equations for the probability of a dependency tree $T$}
\label{nlg3treeequations}
\end{figure*}
In this formulation, the left children are generated independently
from the right children. 
As in NLG2, NLG3 assumes the uniform distribution for the length probabilities 
$Pr(\mbox{\# of left children} = n)$ and 
$Pr(\mbox{\# of right children} = n)$ up to a certain maximum length $M' = 10$.

\subsubsection{Feature Selection}

\begin{table*}
\begin{center}
\begin{tabular}{lp{4.0in}}
Description&Feature $f(ch_i(w),\nlgthreehist) = \dots$\\ \hline
Siblings&1 if $ch_i(w)=?$ and $ch_{i-1}(w) =?$ and $ch_{i-2}(w)=?$ and $dir=?$ and $? \in attr_{w,i}$, 0 otherwise\\
Parent + sibling&1 if $ch_i(w)=?$ and $ch_{i-1}(w) =?$ and $w=?$ and $dir=?$ and $? \in attr_{w,i}$, 0 otherwise\\
Parent + grandparent&1 if $ch_i(w)=?$ and $w =?$ and $par(w)=?$ and $dir=?$ and $? \in attr_{w,i}$, 0 otherwise\\
\end{tabular}
\caption{Features patterns for NLG3. Any occurrence of ``?'' will be instantiated with an actual value from training data.}
\label{nlg3features}
\end{center}
\end{table*}

The feature patterns for NLG3 are shown in Table~\ref{nlg3features}. As before,
the actual features are created by matching the patterns over the training data.
The features in NLG3 have access to syntactic information whereas the features
in NLG2 do not.
Low frequency features involving word $n-$grams tend to be unreliable; the NLG3 system
therefore only uses features which occur $K$ times or more in the training data.
Furthermore, if a feature derived from Table~\ref{nlg3features} 
looks at a particular word $ch_i(w)$ and attribute $a$, we only allow it if $a$
has occurred as a descendent of $ch_i(w)$ in some dependency tree in the training set. 
As an example, this condition allows features that look at $ch_i(w) =$``to'' and \$city-to$\in attr_{w,i}$
but disallows features that look at $ch_i(w) =$``to'' and \$city-fr$\in attr_{w,i}$.

%
% additional filter on features: only choose features f such that the
% attribute they ask about exists as a descendent from the head word
% that feature looks at. 
%

\subsection{Search Procedure}

The idea behind the search procedure for NLG3 is similar to the search procedure for NLG2, 
namely, to explore only a fraction of the possible trees by continually sorting and advancing only
the top $N$ trees at any given point. However, the dependency
trees are not built left-to-right like the word sequences
in NLG2; instead they are built 
from the current head (which is initially the root node)
in the following order:
\begin{enumerate}
\item Predict the next left child (call it $x_l$)
\label{predictleftchildren}
\item If it is $\stopsym$, jump to (\ref{predictrightchildren})
\item Recursively predict children of $x_l$. Resume from (\ref{predictleftchildren})
\item Predict the next right child (call it $x_r$)
\label{predictrightchildren}
\item If it is $\stopsym$, we are done predicting children for the current head
\item Recursively predict children of $x_r$. Resume from (\ref{predictrightchildren})
\end{enumerate}
As before, any incomplete trees that have generated a particular
attribute twice, as well as completed trees that have not generated
a necessary attribute are discarded by the search. 
The search terminates when either $N$ complete trees or $N$ trees of the maximum length $M$ 
are discovered. 
NLG3 chooses the best answer to express the attribute set $A$ as follows: 
\[
nlg_3(A) = \argmax_{T \in T_{nlg3}} Pr(T | A)
\]
where $T_{nlg3}$ are the completed dependency trees that satisfy the conditions
of the NLG3 search described above.

\section{Experiments}

\begin{table*}
\begin{center}
\begin{tabular}{|l|l|llll|l|} \hline
System&Parameters&\% Correct&\% OK&\% Bad&\% No output&\% error reduction\\ 
&&&&&&from NLG1\\ \hline
NLG1&-&84.9&4.9&7.2&3.0&-\\ 
NLG2&N=10,M=30,K=3&88.2&4.7&6.4&0.7&22\\
NLG3&N=5,M=30,K=10&89.9&4.4&5.5&0.2&33\\ \hline
\end{tabular}
\caption{Weighted evaluation of trainable surface generation systems by judge A}
\label{weightedresultsA}
\end{center}
\end{table*}

\begin{table*}
\begin{center}
\begin{tabular}{|l|l|llll|l|} \hline
System&Parameters&\% Correct&\% OK&\% Bad&\% No output&\% error reduction\\ 
&&&&&&from NLG1\\ \hline
NLG1&-&                    81.6& 8.4& 7.0&3.0&-\\ 
NLG2&N=10,M=30,K=3&        86.3& 5.8& 7.2&0.7&26\\
NLG3&N=5,M=30,K=10&  88.4& 4.0& 7.4&0.2&37\\ \hline
\end{tabular}
\caption{Weighted evaluation of trainable surface generation systems by judge B}
\label{weightedresultsB}
\end{center}
\end{table*}

\begin{table*}
\begin{center}
\begin{tabular}{|l|l|llll|l|} \hline
System&Parameters&\% Correct&\% OK&\% Bad&\% No output&\% error reduction\\ 
&&&&&&from NLG1\\ \hline
NLG1&-&48.4&6.8&24.2&20.5&-\\ 
NLG2&N=10,M=30,K=3&64.7&12.1&22.6&0.5&32\\
NLG3&N=5,M=30,K=10&63.1&11.6&23.7&1.6&29\\ \hline
\end{tabular}
\caption{Unweighted evaluation of trainable surface generation systems by judge A}
\label{unweightedresultsA}
\end{center}
\end{table*}

\begin{table*}
\begin{center}
\begin{tabular}{|l|l|llll|l|} \hline
System&Parameters&\% Correct&\% OK&\% Bad&\% No output&\% error reduction\\ 
&&&&&&from NLG1\\ \hline
NLG1&-&                   41.1&8.9&29.5&20.5&-\\ 
NLG2&N=10,M=30,K=3&      62.1&13.7&23.7&0.5&36\\
NLG3&N=5,M=30,K=10&65.3&11.1&22.1&1.6&41\\ \hline
\end{tabular}
\caption{Unweighted evaluation of trainable surface generation systems by judge B}
\label{unweightedresultsB}
\end{center}
\end{table*}

\begin{table*}
\begin{center}
\begin{tt}
\begin{tabular}{ll}
Probability&Generated Text\\ \hline
0.107582&\$time-dep  flights from \$city-fr to \$city-to\\
0.00822441&\$time-dep flights between \$city-fr and \$city-to\\
0.00564712&\$time-dep  flights \$city-fr  to \$city-to\\
0.00343372&flights from \$city-fr   to \$city-to   at \$time-dep\\
0.0012465&\$time-dep  flights from \$city-fr   to  to \$city-to\\   
\end{tabular}
\end{tt}
\caption{Sample output from NLG3. (Dependency tree structures are not shown.) Typical values for attributes: \$time-dep = ``10 a.m.'', \$city-fr = ``New York'', \$city-to = ``Miami''}
\label{nlg3output}
\end{center}
\end{table*}

%  for emacs hiliting..
The training and test sets used to evaluate
NLG1, NLG2 and NLG3 were derived semi-automatically
from a pre-existing annotated corpus of user queries in the 
air travel domain. 
The annotation scheme used a total of 26 attributes to represent flights.
The training set consisted of 6000 templates describing
flights while the test set  consisted of 1946 templates
describing flights. 
All systems used the same training set, 
and were tested on the attribute sets extracted from
the phrases in the test set. 
For example, if the test set contains the template 
``flights to \$city-to leaving at \$time-dep'',
the surface generation systems will be told to generate
a phrase for the attribute set \{ \$city-to, \$time-dep \}.
The output of NLG3 on the attribute set 
\{ \$city-to, \$city-fr, \$time-dep \} is shown in Table~\ref{nlg3output}.

There does not appear to be an objective {\em automatic}
evaluation method\footnote{Measuring word overlap or
edit distance between the system's output and 
a ``reference'' set would be an automatic scoring method.
We believe that such a method does not
accurately measure the correctness or grammaticality
of the text.}
for generated text that correlates
with how an actual person might judge the output. 
Therefore, two judges --- the author and a colleague --- manually
evaluated the output of all three systems. 
Each judge assigned each phrase from each of the three systems one of the 
following rankings:
\begin{description}
\item[Correct:] Perfectly acceptable
\item[OK:] Tense or agreement is wrong, but word choice is correct. 
(These errors could be corrected by post-processing with a morphological analyzer.)
\item[Bad:] Words are missing or extraneous words are present
\item[No Output:] The system failed to produce any output
\end{description}
While there were a total 1946 attribute sets from the 
test examples, the judges only needed to evaluate the
190 {\em unique} attribute sets, e.g., 
the attribute set \{ \$city-fr \$city-to \} occurs
741 times in the test data. 
Subjective evaluation of generation output is not ideal, 
but is arguably superior than
an automatic evaluation that fails to correlate
with human linguistic judgement.

The results of the manual evaluation, as well as the values of the
search and feature selection parameters for all systems, are shown in
Tables~\ref{weightedresultsA}, \ref{weightedresultsB}, \ref{unweightedresultsA}, and \ref{unweightedresultsB}.
(The values for $N$, $M$, and $K$ were determined by manually evaluating the
output of the 4 or 5 most common attribute sets in the training data).
The {\em weighted} results in Tables~\ref{weightedresultsA} and
\ref{weightedresultsB} account for multiple occurrences of attribute
sets, whereas the {\em unweighted} results in
Tables~\ref{unweightedresultsA} and \ref{unweightedresultsB} count each
unique attribute set once, i.e., \{ \$city-fr \$city-to \} is counted
741 times in the weighted results but once in the unweighted results.
Using the weighted results, which represent testing conditions more realistically
than the unweighted results, both judges found an improvement from NLG1 to NLG2, 
and from NLG2 to NLG3.  NLG3 cuts the error rate from NLG1 by at 
least 33\% (counting anything without a
rank of {\bf Correct} as wrong).  NLG2 cuts the error rate by at least
22\% and underperforms NLG3, but requires far less annotation
in its training data.  NLG1 has no chance of generating anything for
3\% of the data --- it fails completely on novel attribute sets.
Using the unweighted results, both judges found an improvement from
NLG1 to NLG2, but, surprisingly, judge A found a slight decrease while
judge B found an increase in accuracy from NLG2 to NLG3.  The
unweighted results show that the baseline NLG1 does well on the common
attribute sets, since it correctly generates only less than 50\% of
the unweighted cases but over 80\% of the weighted cases.

\section{Discussion}

The NLG2 and NLG3 systems automatically attempt to generalize 
from the knowledge  inherent in the training corpus of
templates, so that they can generate templates for novel attribute
sets. There is some additional cost associated with producing the
syntactic dependency annotation necessary for NLG3, but virtually
no additional cost is associated with NLG2, beyond collecting the
data itself and identifying the attributes.

The trainable surface NLG systems in this paper differ from
grammar-based systems in how they determine the
attribute ordering and lexical choice.
NLG2 and NLG3 automatically determine
attribute ordering 
by simultaneously searching multiple orderings.  
In grammar-based approaches, such preferences need to be manually encoded.  
NLG2 and NLG3 solve the lexical choice problem 
by learning the words (via features in the maximum entropy probability model) 
that correlate with a given attribute and local context, 
whereas \cite{lexchoice97} uses a rule-based approach to decide
the word choice.

While trainable approaches avoid the expense of crafting a grammar to
determine attribute ordering and lexical choice, they are less
accurate than grammar-based approaches.  For short phrases,
accuracy is typically 100\% with grammar-based approaches
since the grammar writer can either correct or add a rule to generate
the phrase of interest once an error is detected. Whereas with NLG2
and NLG3, one can tune the feature patterns, search parameters, and
training data itself, but there is no guarantee that the tuning will
result in 100\% generation accuracy.

Our approach differs from the corpus-based surface generation
approaches of \cite{langkilde98} and \cite{maxent:mt}. 
\cite{langkilde98} maps from semantics to words with a concept
ontology, grammar, and lexicon, and ranks the resulting word
lattice with corpus-based statistics, whereas NLG2 and NLG3 
automatically learn the mapping from semantics to words from a corpus.
\cite{maxent:mt} describes a statistical machine translation approach
that generates text in the target language directly from the source
text. NLG2 and NLG3 are also statistical learning approaches but 
generate from an actual semantic representation.
This comparison suggests that statistical MT systems could also
generate text from an ``interlingua'', in a way similar to that
of knowledge-based translation systems.

We suspect that our statistical generation approach should perform
accurately in domains of similar complexity to air travel.
In the air travel domain,  the length of a phrase fragment to describe an attribute is usually only a few words.
Domains which require complex and lengthy phrase fragments to describe a single attribute
will be more challenging to model with features that only look at word $n$-grams for $n \in \{ 2,3 \}$.
Domains in which there is greater ambiguity in word choice will require a more 
thorough search, i.e., a larger value of $N$, at the expense of CPU time and memory.
Most importantly, the semantic annotation scheme for air travel has the property that
it is  both rich enough to accurately represent meaning in the domain, but simple 
enough to yield useful corpus statistics. 
Our approach may not scale to domains, such as freely occurring newspaper text, in which the semantic
annotation schemes do not have this property.

Our current approach has the limitation that it ignores
the values of attributes, even 
though they might strongly influence the word order and word choice. 
This limitation can be overcome by using features on values, so 
that NLG2 and NLG3 might discover --- to use a hypothetical example ---  that 
``flights leaving \$city-fr'' is preferred over ``flights from \$city-fr'' 
when \$city-fr is a particular value, such as ``Miami''.

\section{Conclusions}

This paper presents  the first systems (known to the author) that use a statistical learning 
approach to produce natural language text directly from a semantic representation.
Information to solve the attribute ordering and lexical choice problems---which 
would normally be specified in a large hand-written grammar---
is automatically collected from data with a few feature patterns, 
and is combined via the maximum entropy framework. 
NLG2 shows that using just local $n$-gram information can outperform
the baseline, and NLG3 shows that using syntactic information 
can further improve generation accuracy. 
We conjecture that NLG2 and NLG3 should work in other domains
which have a complexity similar to air travel, as well as 
available annotated data.

\section{Acknowledgements}

The author thanks Scott McCarley for serving as the second judge, and Scott Axelrod, Kishore Papineni, and Todd Ward for their helpful comments on this work. 
This work was supported in part by DARPA Contract \# MDA972-97-C-0012. 

\bibliographystyle{acl}

\bibliography{refs}

\end{document}